%% file: main.tex
\documentclass[10pt,twocolumn,letterpaper]{article}

\usepackage{cvpr}              

\input{preamble}

\definecolor{cvprblue}{rgb}{0.21,0.49,0.74}
\usepackage[pagebackref,breaklinks,colorlinks,citecolor=cvprblue]{hyperref}

\usepackage{indentfirst}
\usepackage{graphicx}
\usepackage{tikz}
\usetikzlibrary{shapes, arrows, positioning}
\usetikzlibrary{fit}
\usepackage{pgfplots}
\pgfplotsset{compat=1.18}
\usepackage{algorithm}
\usepackage{algpseudocode}
\MakeRobust{\Call}

\def\paperID{1764} 
\def\confName{CVPR}
\def\confYear{2024}

\title{Reconstructing the Invisible: Video Frame Restoration through Siamese Masked Conditional Variational Autoencoder}

\author{Yongchen Zhou\\
LIRA Center, Lancaster University\\
Lancaster, England\\
{\tt\small y.zhou52@lancaster.ac.uk}
\and
Richard Jiang\\
LIRA Center, Lancaster University\\
Lancaster, England\\
{\tt\small r.jiang2@lancaster.ac.uk}
}

\begin{document}
\maketitle
\input{sec/0_abstract}    
\input{sec/1_intro}
\input{sec/2_relat}
\input{sec/3_metho}
\input{sec/4_exper}
\input{sec/5_discu}
\input{sec/6_concl}


{
    \small
    \bibliographystyle{ieeenat_fullname}
    \bibliography{main}
}

\end{document}

%% file: preamble.tex
\usepackage[dvipsnames]{xcolor}


%% file: sec/0_abstract.tex
\begin{abstract}

In the domain of computer vision, the restoration of missing information in video frames is a critical challenge, particularly in applications such as autonomous driving and surveillance systems. This paper introduces the Siamese Masked Conditional Variational Autoencoder (SiamMCVAE), leveraging a siamese architecture with twin encoders based on vision transformers. This innovative design enhances the model's ability to comprehend lost content by capturing intrinsic similarities between paired frames. SiamMCVAE proficiently reconstructs missing elements in masked frames, effectively addressing issues arising from camera malfunctions through variational inferences. Experimental results robustly demonstrate the model's effectiveness in restoring missing information, thus enhancing the resilience of computer vision systems. The incorporation of Siamese Vision Transformer (SiamViT) encoders in SiamMCVAE exemplifies promising potential for addressing real-world challenges in computer vision, reinforcing the adaptability of autonomous systems in dynamic environments.
\end{abstract}

%% file: sec/1_intro.tex
\section{Introduction}
In the dynamic world of computer vision, where the lens of artificial intelligence gazes upon the visual landscape, a singular challenge has continued to captivate the imaginations of researchers and engineers alike. This challenge lies at the intersection of technology and the human experience—a quest to restore what has been lost \cite{he2022masked}, to unveil the unseen, and to breathe life into the incomplete. In a world fueled by the relentless pursuit of innovation, the restoration of missing information within video frames stands as a formidable testament to the artistry of visual intelligence \cite{liang2022recurrent}.

In recent years, developments in the field of deep learning have witnessed a growing movement towards the integration of methodologies to address a wide array of challenges, encompassing language \cite{kowsari2019text}, vision \cite{chen2023symbolic, karras2019style}, speech \cite{zhang2020pushing}, and various other domains. The adaptation of Transformer architectures \cite{vaswani2017attention}, initially prevalent in natural language processing, has found successful integration into the realm of computer vision \cite{dosovitskiy2020image}. The landscape of predictive learning methods has witnessed an intriguing evolution, driven by the transformative potential of masked language modeling \cite{devlin2018bert, brown2020language} and its visual counterpart, masked visual modeling (MVM) \cite{he2022masked, bao2021beit, xie2022simmim}.

\begin{figure*}
  \centering
  \resizebox{\textwidth}{!}{
    \begin{tikzpicture}[>=stealth]
      \tikzstyle{block} = [rectangle, draw, text width=5.5em, text centered, rounded corners, minimum height=4em]
      \tikzstyle{operator} = [circle, draw, minimum size=2em, inner sep=1pt, fill=yellow!30]
      \tikzstyle{latent} = [circle, draw, minimum size=2em, inner sep=1pt, fill=gray!30]
      \tikzstyle{line} = [draw, -latex']
      \tikzstyle{rect} = [rectangle, draw=none, minimum width=1em, minimum height=1em, rounded corners=0.1em]

      \node [inner sep=0pt] (frame1) {\includegraphics[width=6em]{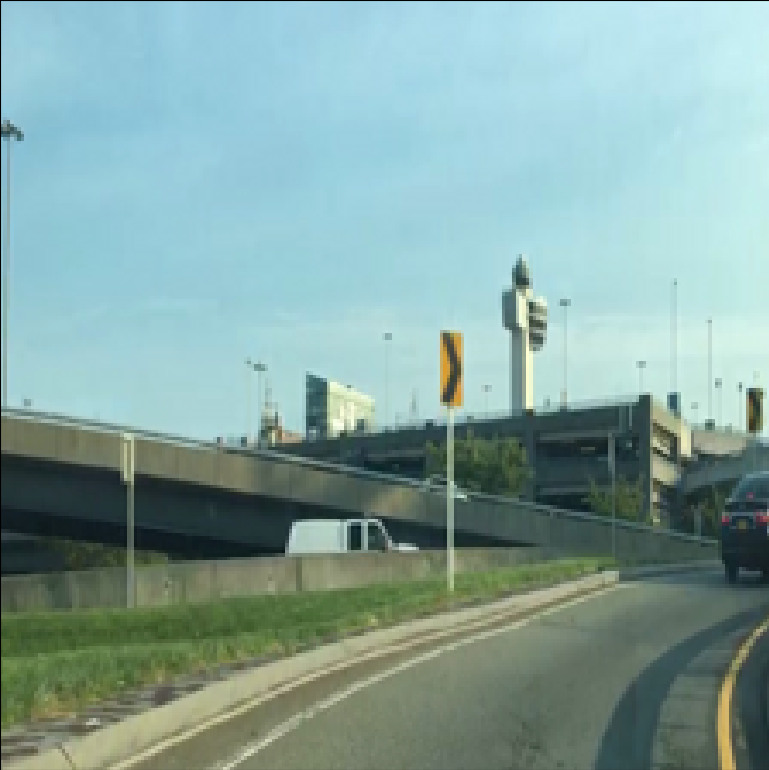}};
      \node [orange, above=0.75em of frame1] {frame 1};
      \node [inner sep=0pt, right=2em of frame1] (patchified1) {\includegraphics[width=6em]{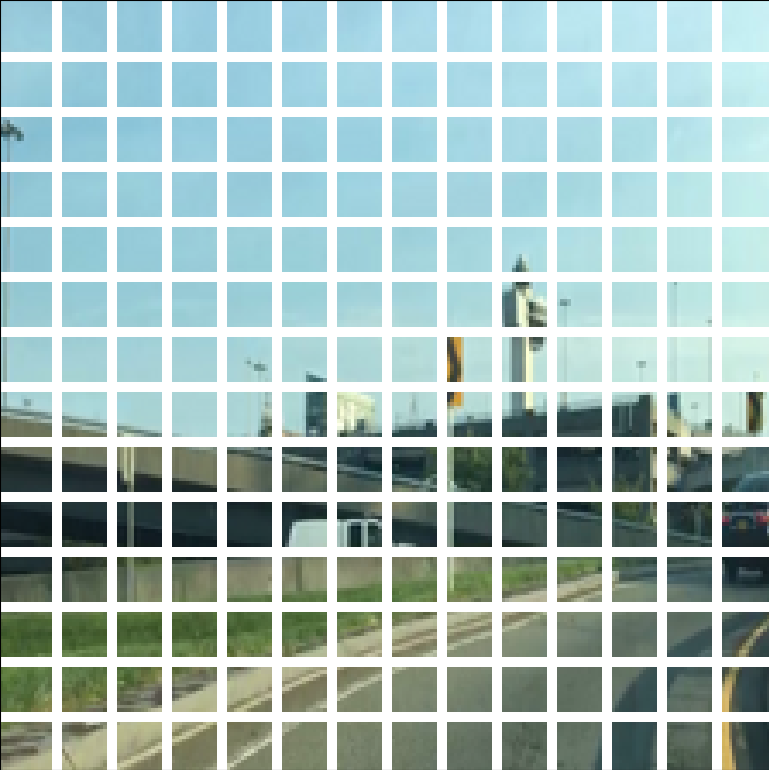}};
      \node [orange, above=0.75em of patchified1] {patchify};

      \node [right=2em of patchified1] (vdots1) {$\vdots$};
      \node [rect, fill=orange!60, above=-0.5em of vdots1] (rect13) {};
      \node [rect, fill=orange!60, above=0.1em of rect13] (rect12) {};
      \node [rect, fill=orange!60, above=0.1em of rect12] (rect11) {};
      \node [rect, fill=orange!60, below=0.1em of vdots1] (rect13) {};
      \node [rect, fill=orange!60, below=0.1em of rect13] (rect14) {};

      \node [block, right=2em of vdots1, fill=red!30] (SiamViT1) {SiamViT};

      \node [right=2em of SiamViT1] (vdots3) {$\vdots$};
      \node [rect, fill=red!60, above=-0.5em of vdots3] (rect33) {};
      \node [rect, fill=red!60, above=0.1em of rect33] (rect32) {};
      \node [rect, fill=red!60, above=0.1em of rect32] (rect31) {};
      \node [rect, fill=red!60, below=0.1em of vdots3] (rect33) {};
      \node [rect, fill=red!60, below=0.1em of rect33] (rect34) {};

      \node [inner sep=0pt, below=4em of frame1] (frame2) {\includegraphics[width=6em]{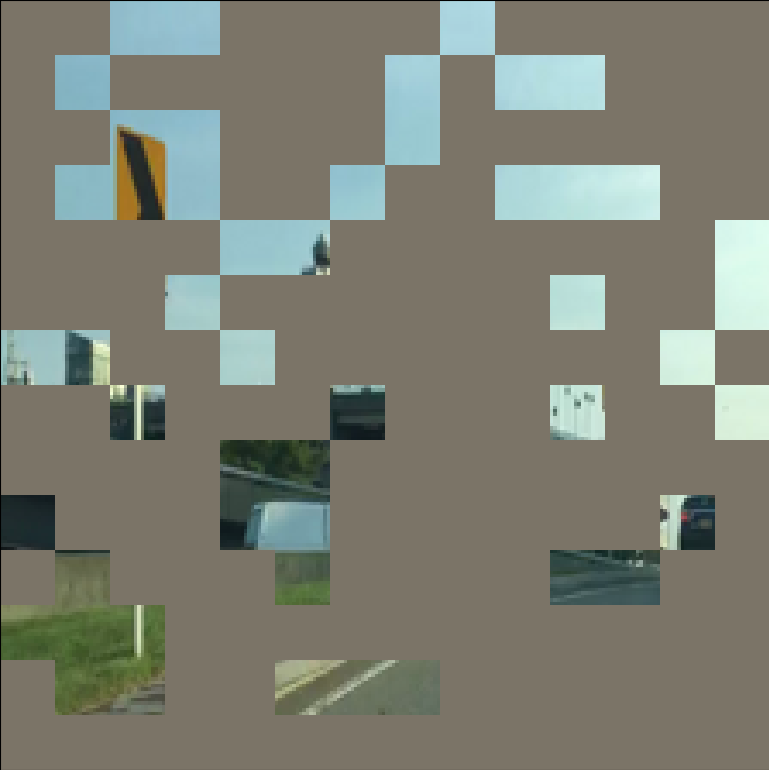}};
      \node [violet, below=0.75em of frame2] {frame 2};
      \node [inner sep=0pt, right=2em of frame2] (patchified2) {\includegraphics[width=6em]{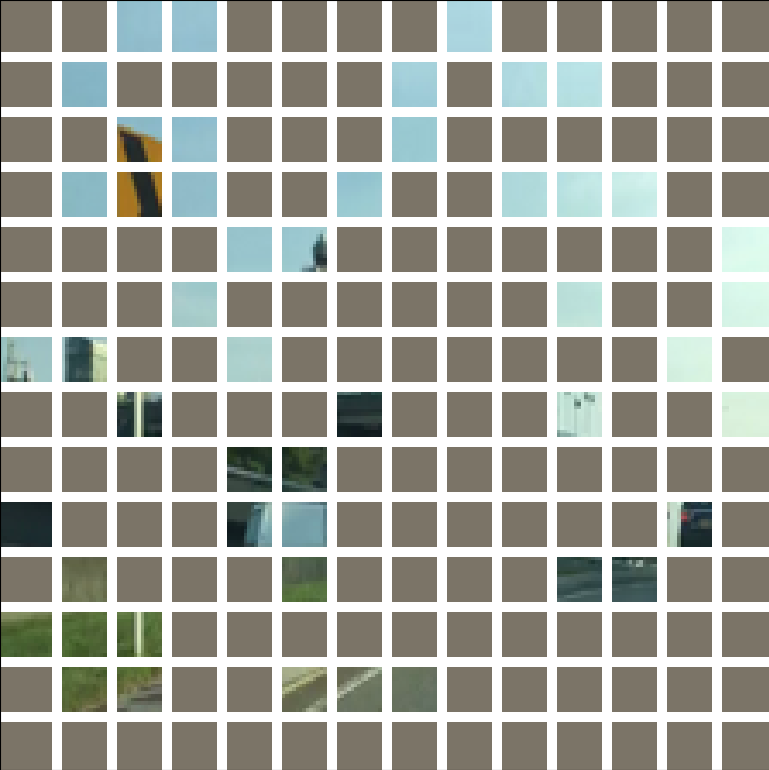}};
      \node [violet, below=0.75em of patchified2] {patchify};
      
      \node [right=2em of patchified2] (vdots2) {$\vdots$};
      \node [rect, fill=violet!60, above=-0.5em of vdots2] (rect23) {};
      \node [rect, fill=violet!60, above=0.1em of rect23] (rect22) {};
      \node [rect, fill=violet!60, below=0.1em of vdots2] (rect24) {};

      \node [block, right=2em of vdots2, fill=red!30] (SiamViT2) {SiamViT};
      \node [right=2em of SiamViT2] (vdots4) {$\vdots$};
      \node [rect, fill=blue!60, above=-0.5em of vdots4] (rect43) {};
      \node [rect, fill=blue!60, above=0.1em of rect43] (rect42) {};
      \node [rect, fill=blue!60, below=0.1em of vdots4] (rect44) {};

      \node [block, below left=5em and 2em of rect44, fill=brown!20] (convert1) {Convert};
      \node [above=2em of convert1] (Nplus11) {$N+1$};
      
      \node [below=4em of convert1] (vdots5) {$\vdots$};
      \node [rect, fill=blue!60, above=-0.5em of vdots5] (rect53) {};
      \node [rect, fill=gray!60, above=0.1em of rect53] (rect52) {};
      \node [rect, fill=gray!60, above=0.1em of rect52] (rect51) {};
      \node [rect, fill=gray!60, below=0.1em of vdots5] (rect54) {};
      \node [rect, fill=gray!60, below=0.1em of rect54] (rect55) {};

      \node [latent, left=5em of convert1] (P1) {$\mathcal{P}$};
      \node [latent, below=4em of P1] (PC1) {$\mathcal{P}^\complement$};
      \node [block, below=4em of PC1, fill=brown!20] (convert2) {Convert};
      \node [below=2em of convert2] (Nplus12) {$N+1$};

      \node [left=3em of convert2] (vdots6) {$\vdots$};
      \node [rect, fill=olive!60, above=-0.5em of vdots6] (rect63) {};
      \node [rect, fill=olive!60, above=0.1em of rect63] (rect62) {};
      \node [rect, fill=olive!60, below=0.1em of vdots6] (rect64) {};

      \node [block, above=2em of rect62, fill=cyan!20] (repeat) {Repeat};
      \node [left=2em of repeat] (N1) {$\lvert \mathcal{P} \lvert$};
      \node [rect, fill=olive!60, above=2em of repeat] (token) {};
      \node [olive, above=0.5em of token] {mask token};

      \node [right=2em of convert2] (vdots7) {$\vdots$};
      \node [rect, fill=gray!60, above=-0.5em of vdots7] (rect73) {};
      \node [rect, fill=olive!60, above=0.1em of rect73] (rect72) {};
      \node [rect, fill=olive!60, above=0.1em of rect72] (rect71) {};
      \node [rect, fill=olive!60, below=0.1em of vdots7] (rect74) {};
      \node [rect, fill=olive!60, below=0.1em of rect74] (rect75) {};

      \node [operator, right=2em of vdots7] (plus1) {+};

      \node [right=2em of plus1] (vdots8) {$\vdots$};
      \node [rect, fill=blue!60, above=-0.5em of vdots8] (rect83) {};
      \node [rect, fill=olive!60, above=0.1em of rect83] (rect82) {};
      \node [rect, fill=olive!60, above=0.1em of rect82] (rect81) {};
      \node [rect, fill=olive!60, below=0.1em of vdots8] (rect84) {};
      \node [rect, fill=olive!60, below=0.1em of rect84] (rect85) {};

      \node [block, above right=0em and 3em of vdots4, fill=green!20] (projection1) {Projection};
      \node [block, below=2em of projection1, fill=violet!30] (sample) {$\mathcal{MN}(0, \mathbf{I}, \mathbf{I})$};
      \node [latent, above right=-0.5em and 2em of projection1] (M) {$\mathbf{M}$};
      \node [latent, below right=-0.5em and 2em of projection1] (S) {$\mathbf{S}$};
      \node [latent, right=1.7em of sample] (E) {$\mathbf{E}$};
      \node [operator, right=2em of S] (odot) {$\odot$};
      \node [operator, right=5em of M] (plus2) {+};
      \node [latent, right=2em of plus2] (Z) {$\mathbf{Z}$};
      
      \node [block, right=19.9em of vdots3, fill=green!20] (projection2) {Projection};
      \node [block, right=2em of projection2, fill=blue!20] (ViT) {ViT};

      \node [below=7em of ViT] (vdots9) {$\vdots$};
      \node [rect, fill=cyan!60, above=-0.5em of vdots9] (rect93) {};
      \node [rect, fill=cyan!60, above=0.1em of rect93] (rect92) {};
      \node [rect, fill=cyan!60, above=0.1em of rect92] (rect91) {};
      \node [rect, fill=cyan!60, below=0.1em of vdots9] (rect94) {};
      \node [rect, fill=cyan!60, below=0.1em of rect94] (rect95) {};

      \node [block, below left=6em and 7em of vdots9, fill=brown!20] (convert3) {Convert};
      \node [above=2em of convert3] (N2) {$N$};
      \node [latent, left=5em of convert3] (PC2) {$\mathcal{P}^\complement$};
      \node [latent, below=4em of PC2] (P2) {$\mathcal{P}$};

      \node [below=4em of convert3] (vdots10) {$\vdots$};
      \node [rect, fill=gray!60, above=-0.5em of vdots10] (rect103) {};
      \node [rect, fill=cyan!60, above=0.1em of rect103] (rect102) {};
      \node [rect, fill=cyan!60, above=0.1em of rect102] (rect101) {};
      \node [rect, fill=cyan!60, below=0.1em of vdots10] (rect104) {};
      \node [rect, fill=cyan!60, below=0.1em of rect104] (rect105) {};

      \node [block, below=10em of PC2, fill=brown!20] (convert4) {Convert};
      \node [below=2em of convert4] (N3) {$N$};

      \node [left=2em of convert4] (vdots11) {$\vdots$};
      \node [rect, fill=violet!60, above=-0.5em of vdots11] (rect113) {};
      \node [rect, fill=violet!60, above=0.1em of rect113] (rect112) {};
      \node [rect, fill=violet!60, below=0.1em of vdots11] (rect114) {};

      \node [right=2em of convert4] (vdots12) {$\vdots$};
      \node [rect, fill=violet!60, above=-0.5em of vdots12] (rect123) {};
      \node [rect, fill=gray!60, above=0.1em of rect123] (rect122) {};
      \node [rect, fill=gray!60, above=0.1em of rect122] (rect121) {};
      \node [rect, fill=gray!60, below=0.1em of vdots12] (rect124) {};
      \node [rect, fill=gray!60, below=0.1em of rect124] (rect125) {};

      \node [operator, right=2em of vdots12] (plus3) {+};
      
      \node [right=2em of plus3] (vdots13) {$\vdots$};
      \node [rect, fill=violet!60, above=-0.5em of vdots13] (rect133) {};
      \node [rect, fill=cyan!60, above=0.1em of rect133] (rect132) {};
      \node [rect, fill=cyan!60, above=0.1em of rect132] (rect131) {};
      \node [rect, fill=cyan!60, below=0.1em of vdots13] (rect134) {};
      \node [rect, fill=cyan!60, below=0.1em of rect134] (rect135) {};

      \node [inner sep=0pt, right=3em of vdots13] (target) {\includegraphics[width=6em]{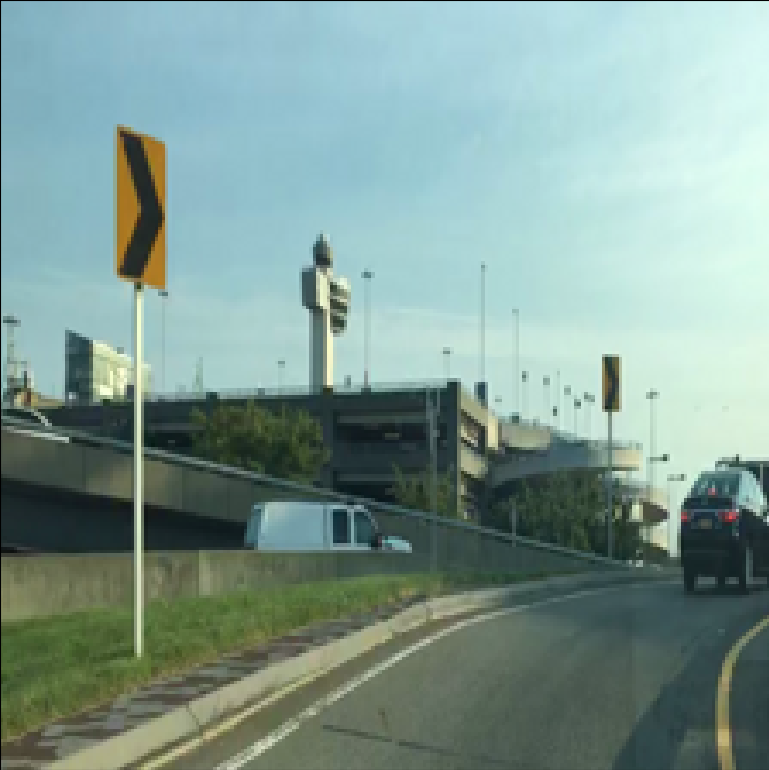}};
      \node [cyan, below=0.75em of target] {target};
      
      \path [line] (frame1) -- (patchified1);
      \path [line] (patchified1) -- (vdots1);
      \path [line] (vdots1) -- (SiamViT1);
      \path [line] (SiamViT1) -- (vdots3);
      \path [line] (vdots3) -| (projection1);
      \path [line] (vdots3) -- (projection2);

      \path [line] (frame2) -- (patchified2);
      \path [line] (patchified2) -- (vdots2);
      \path [line] (vdots2) -- (SiamViT2);
      \path [line] (SiamViT2) -- (vdots4);

      \path [line] (rect44) |- (convert1);
      \path [line] (convert1) -- (rect51);
      \path [line] (rect55) -- (plus1);

      \path [line] (P1) -- (PC1);
      \path [line] (Nplus11) -- (convert1);
      \path [line] (P1) -- (convert1);
      \path [line] (PC1) -- (convert2);
      \path [line] (vdots6) -- (convert2);

      \path [line] (N1) -- (repeat);
      \path [line] (token) -- (repeat);
      \path [line] (repeat) -- (rect62);

      \path [line] (Nplus12) -- (convert2);
      \path [line] (convert2) -- (vdots7);
      \path [line] (vdots7) -- (plus1);
      \path [line] (plus1) -- (vdots8);
      \path [line] (vdots8) -- ++(1.5,0) |- (projection1);

      \path [line] (projection1) -- (M);
      \path [line] (projection1) -- (S);
      \path [line] (sample) -- (E);
      \path [line] (S) -- (odot);
      \path [line] (E) -| (odot);
      \path [line] (M) -- (plus2);
      \path [line] (odot) -| (plus2);
      \path [line] (plus2) -- (Z);
      \path [line] (Z) -- (projection2);
      \path [line] (projection2) -- (ViT);

      \path [line] (ViT) -- (rect91);
      \path [line] (N2) -- (convert3);
      \path [line] (rect95) |- (convert3);

      \path [line] (P2) -- (PC2);
      \path [line] (PC2) -- (convert3);
      \path [line] (convert3) -- (rect101);
      \path [line] (rect105) -- (plus3);
      \path [line] (N3) -- (convert4);
      \path [line] (P2) -- (convert4);
      \path [line] (vdots11) -- (convert4);
      \path [line] (convert4) -- (vdots12);
      \path [line] (vdots12) -- (plus3);
      \path [line] (plus3) -- (vdots13);
      \path [line] (vdots13) -- (target);
      
    \end{tikzpicture}
  }
  \caption{\textbf{Our SiamMCVAE architecture.} The foundational framework of our SiamMCVAE is meticulously crafted to address the intricate challenges posed by missing information in video frames. Embracing a siamese architecture, our model synergistically integrates twin encoders equipped with vision transformers. This innovative design augments the model's ability to discern and reconstruct missing content by capturing inherent similarities between paired frames. The siamese encoder configuration, coupled with the transformative power of vision transformers, empowers SiamMCVAE to proficiently reconstruct missing elements within masked frames. The intricacies of our architecture extend further with the incorporation of variational principles, elevating the model's capacity to generate diverse and meaningful representations.}
  \label{fig:SiamMCVAE}
\end{figure*}
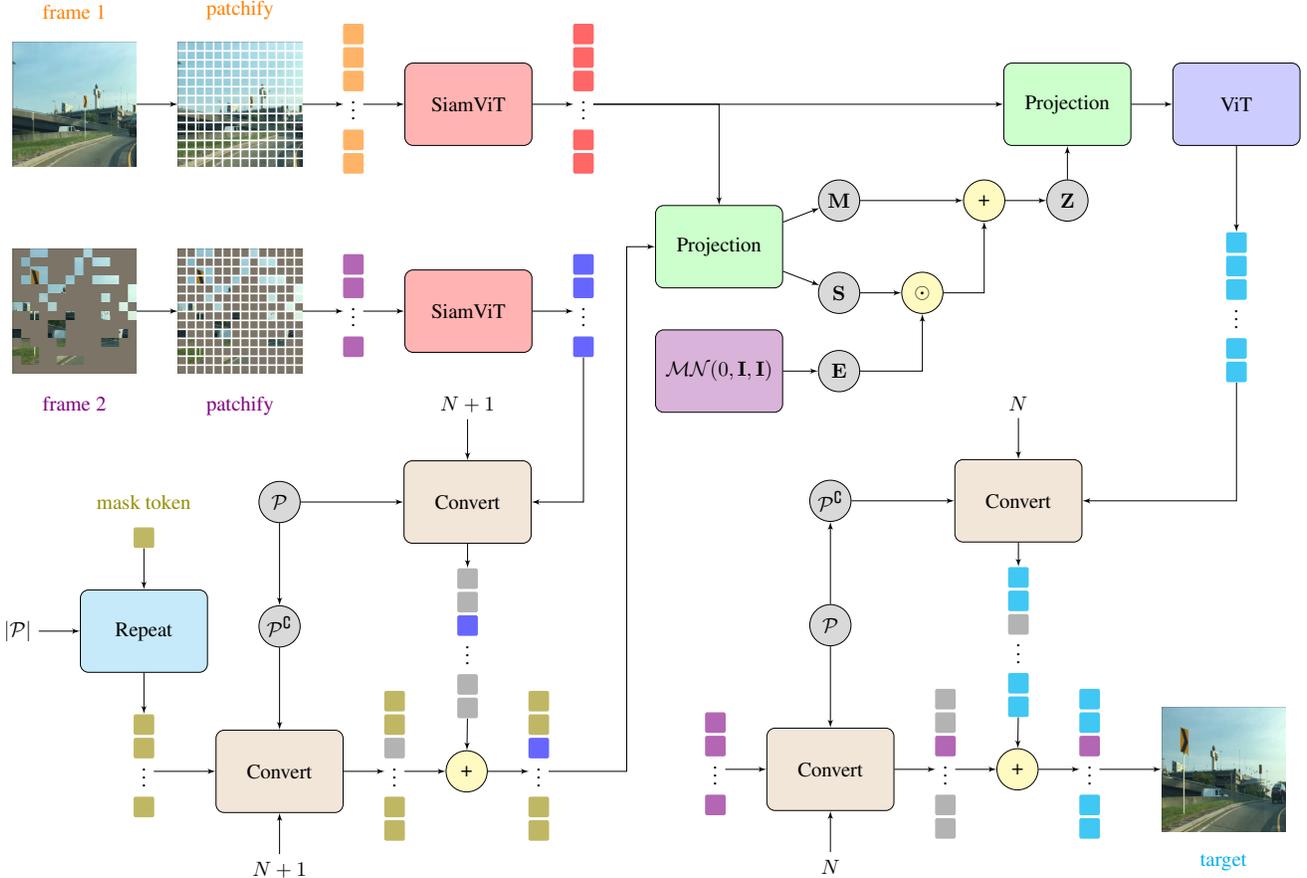

This paper confronts the formidable challenge of restoring large-scale missing information within video frames, introducing a groundbreaking solution that harnesses the latest advancements in machine learning and computer vision. Our model, SiamMCVAE, illustrated in \Cref{fig:SiamMCVAE}, draws inspiration from the Conditional Variational Autoencoder (CVAE) \cite{sohn2015learning}, ushering in a significant breakthrough in the realm of restoration capability. While siamese networks \cite{sohn2015learning} have conventionally found applications in classification and comparison tasks \cite{wu2018unsupervised, he2020momentum, caron2021emerging, chen2020simple}, our work extends this architecture to the generative domain, introducing a novel dimension to its utilization.

The existing Masked Autoencoders (MAE) \cite{he2022masked} and their extensions \cite{feichtenhofer2022masked, tong2022videomae} demonstrate proficiency in restoring large-scale missing information. However, these models lack comprehensive evaluations specifically focused on image restoration. To address this void, our meticulous evaluation uniquely scrutinizes the performance of these models, placing a distinct emphasis on their efficacy in the context of image restoration. Through this investigation, we unveil SiamMCVAE's unparalleled advantages over them. Notably, its exceptional capability to excel in reconstructing images, even in scenarios characterized by extensive missing information, establishes it as a pioneering solution in the field.

What sets our model apart is its remarkable ability to excel in restoring information under challenging conditions. SiamMCVAE, with its unique capacity to learn correspondences and reconstruct lost patches within video frames, positions itself as a pioneer in the field of computer vision. Our extensive experiments and results unequivocally demonstrate the superiority of our model in comparison to existing methods, showcasing its potential to revolutionize the field.

%% file: sec/2_relat.tex
\section{Related Work}
\textbf{Autoencoder.} Autoencoders, integral to unsupervised learning, aim to distill intricate data representations and excel in reconstructing the original data from this condensed form \cite{schmidhuber2015deep}. This architecture encompasses an encoder, responsible for mapping inputs to a latent representation, and a decoder, tasked with reconstructing the input. Well-established instances of autoencoders include Principal Component Analysis (PCA) \cite{karush1939minima} and k-means \cite{hinton1993autoencoders}. In this domain, Denoising Autoencoders (DAE) \cite{vincent2008extracting} represent a specialized class deliberately introducing corruption to input signals, striving to learn the reconstruction of the original, uncorrupted signal. Moreover, various methods can be conceptualized as generalized DAEs employing diverse corruption techniques, such as masking pixels \cite{vincent2010stacked,pathak2016context, chen2020rewon}, or removing color channels \cite{zhang2016colorful}. Our work is specifically tailored to restoring frames where information in a substantial proportion of patches has been lost.

\textbf{Variational inference.} Variational inference \cite{blei2017variational} is a powerful framework in probabilistic modeling that enables the approximation of complex posterior distributions. It is particularly valuable when dealing with intractable probabilistic models. The primary goal of variational inference is to find an approximate distribution, usually denoted as $q(\mathbf{z})$, that closely approximates the true posterior distribution, $p(\mathbf{z} | \mathbf{x})$, where $\mathbf{z}$ represents latent variables and $\mathbf{x}$ represents observed data.

The core idea of variational inference is to transform the posterior inference problem into an optimization problem. By minimizing the Kullback-Leibler (KL) divergence \cite{kullback1951information} between the approximate distribution $q(\mathbf{z})$ and the true posterior $p(\mathbf{z} | \mathbf{x})$, we can find the best approximation:

\begin{equation}
  q^*(\mathbf{z}) = \underset{q(\mathbf{z})}{\mathrm{argmin}} \, D_\mathrm{KL}(q(\mathbf{z}) \| p(\mathbf{z} | \mathbf{x})).
\end{equation}

Here, the KL divergence measures the information lost when using the approximate distribution instead of the true posterior. The optimal approximation, $q^*(\mathbf{z})$, provides a trade-off between being expressive enough to capture the true posterior and being computationally tractable.

Variational inference has found extensive applications in machine learning, including in the training of Variational Autoencoders (VAE) \cite{kingma2013auto}, Conditional Variational Autoencoders (CVAE) \cite{sohn2015learning}, and other generative models. It enables the efficient learning of complex probabilistic models and has become an essential tool in the field of deep learning.

\textbf{Siamese networks.} Siamese networks have emerged as a significant architectural paradigm in the field of computer vision and machine learning \cite{bromley1993signature}. Their unique ability to compare entities by means of weight-sharing neural networks has found broad application across diverse domains, and has been extensively featured in the contrastive learning approaches \cite{wu2018unsupervised, he2020momentum, caron2021emerging, chen2020simple}, showcasing its versatility and efficacy in capturing complex relationships.

In our work, we transcend the conventional boundaries of siamese networks by venturing into the generative domain, thereby introducing a novel dimension to its application. This expansion unlocks new possibilities for leveraging siamese architectures in tasks related to generative modeling and content restoration.

\textbf{Data restoration.} Traditional denoising methods \cite{chen2022simple, zamir2022restormer} demonstrate proficiency in managing noisy images. However, their efficacy experiences a considerable decline when faced with scenarios involving substantial missing regions. In recent years, MAE \cite{he2022masked} and its variants \cite{feichtenhofer2022masked, tong2022videomae} have surfaced as leading methodologies for addressing masked scenarios in video frames. These models employ sophisticated representations to reconstruct missing information.

Our work builds upon these foundations, introducing the SiamMCVAE model, which combines the strengths of Siamese architectures and Vision Transformers for enhanced data restoration. Unlike some existing approaches that might prioritize specific aspects of masked scenarios, our model takes a holistic approach, focusing on comprehensive image restoration, even in situations with large-scale missing information. This distinctive emphasis positions SiamMCVAE as a robust and versatile solution in the landscape of data restoration.

%% file: sec/3_metho.tex
\section{Method}
In this section, we undertake an in-depth exploration of the fundamental components comprising our SiamMCVAE model. Our method amalgamates cutting-edge technologies in computer vision and machine learning, underpinned by the principles of SiamViT and variational inference \cite{blei2017variational}. This synthesis of innovative concepts culminates in a comprehensive solution designed to tackle the intricate challenges posed by missing information in video frames, thus bolstering the efficacy of computer vision systems operating in rapidly evolving scenarios.

To provide a concrete understanding of the inner workings of SiamMCVAE, we present the forward propagation function outlined in \Cref{alg:SiamMCVAE}. This algorithm serves as the blueprint for the model's forward propagation, elucidating the sequential steps involved in processing input data and generating meaningful output representations. The subsequent sections delve into a detailed discussion of the various components of SiamMCVAE, shedding light on their roles and contributions to the overall framework.

\begin{algorithm}
  \caption{Forward Propagation of SiamMCVAE}
  \label{alg:SiamMCVAE}
  
  \begin{algorithmic}
    \Function{Convert}{$\mathbf{X}, \mathcal{P}, N$}
      \State $M, D \gets$ \Call{rows}{$\mathbf{X}$}$,$ \Call{cols}{$\mathbf{X}$}
      \For{$i \gets 1$ to $N$}
        \If{$i - 1 \in \mathcal{P}$}
          \State $\mathbf{y}_{i} \gets \mathbf{0}$
        \Else
          \If{$N \leq M$}
            \State $k \gets i + M - N$
          \Else
            \State $k \gets i - \lvert \mathcal{P} \cap \{1, 2, \ldots, i-1\} \lvert$
          \EndIf
          \State $\mathbf{y}_{i} \gets (\mathbf{X}_{k1}, \mathbf{X}_{k2}, \ldots, \mathbf{X}_{kD})^\mathsf{T}$
        \EndIf
      \EndFor
      \State \Return $[\mathbf{y}_1, \mathbf{y}_2, \ldots, \mathbf{y}_N]^\mathsf{T}$
    \EndFunction

    \Function{SiamMCVAE}{$\mathbf{X}_1, \mathbf{X}_2, \mathcal{P}$}
      \State $\mathbf{X}_1 \gets$ \Call{Patchify}{$\mathbf{A}_1, \{1, 2, \ldots, N \}$}
      \State $\mathbf{X}_2 \gets$ \Call{Patchify}{$\mathbf{A}_2, \mathcal{P}^\complement$}
      \State $\mathbf{U}_1 \gets$ \Call{SiamViT}{$\mathbf{X}_1$}
      \State $\mathbf{U}_2 \gets$ \Call{SiamViT}{$\mathbf{X}_2$}
      \State $\mathbf{T} \gets$ \Call{Repeat}{$\mathbf{t}^\mathsf{T}, \lvert \mathcal{P}^\complement \lvert$}
      \State $\mathbf{U} \gets [\mathbf{U}_1,$ \Call{Convert}{$\mathbf{U}_2, \mathcal{P}, N+1$} $+$ \Call{Convert}{\newline $\mathbf{T}, \mathcal{P}, N+1$} $]$
      \State $\mathbf{Z}, \mathbf{M}, \mathbf{S} \gets$ \Call{Reparametrize}{$\mathbf{U}$}
      \State $\mathbf{O} \gets$ \Call{ViT}{$[\mathbf{Z}, \mathbf{U}_1]$}
      \State $\mathbf{G} \gets$ \Call{Convert}{$[\mathbf{0}^\mathsf{T}; \mathbf{X}_2], \mathcal{P}, N$} $+$ \Call{Convert}{$\mathbf{O}, \newline \mathcal{P}^\complement, N$}
      \State \Return $\mathbf{G}, \mathbf{M}, \mathbf{S}$
    \EndFunction
  \end{algorithmic}
\end{algorithm}

\textbf{Siamese encoder.} The encoding process commences with the patchification of each video frame pair. We perform a transformation on the images $\mathbf{A}_1$ and $\mathbf{A}_2 \in \mathbb{R}^{H \times W \times C}$ by converting them into sequences of flattened 2D patches, denoted as $\mathbf{X}_1$ and $\mathbf{X}_2 \in \mathbb{R}^{N \times (P^2 \cdot C)}$, where $H \times W$ represents the resolution of the original images, $C$ is the number of channels, $P \times P$ denotes the resolution of each image patch, and $N = \frac{HW}{P^2}$ signifies the resulting number of patches. Crafted explicitly for processing pairs of video frames, the SiamViT adeptly manages paired data with the utilization of two weight-sharing vanilla Vision Transformers (ViT) \cite{dosovitskiy2020image}. This independent processing of video frame pairs involves one intact frame and another subjected to masking.

The SiamViT architecture embodies a sophisticated design, featuring a cascade of interleaved Multiheaded Self-Attention (MSA) \cite{vaswani2017attention} and Multilayer Perceptron (MLP) \cite{tolstikhin2021mlp} blocks. The MSA employs adaptive attention kernel, dynamically selecting the most optimal implementation based on the characteristics of the input data. The available implementations include Standard Attention \cite{vaswani2017attention}, Flash Attention \cite{dao2022flashattention}, and Memory-Efficient Attention \cite{jeevan2022resource}. The choice among these implementations is made to maximize efficiency and performance. A strategic application of Layer Normalization (LN) precedes each block, augmenting the stability and efficiency of the model. Further bolstering the network's expressiveness, residual connections are strategically integrated after each block, contributing to seamless information flow and facilitating effective gradient propagation \cite{wang2019learning, baevski2018adaptive}. Mathematically, the SiamViT operations can be represented as follows:
\begin{gather}
  \mathbf{Y}_{i,0} = [\mathbf{c}, \mathbf{W}_\mathrm{e} \mathbf{X}_i^\mathsf{T} + \mathbf{B}_\mathrm{e}]^\mathsf{T} + \mathbf{P}_\mathrm{e}, \\
    \mathbf{Y}'_{i,l} = \mathrm{MSA}_l(\mathrm{LN}(\mathbf{Y}_{i,l-1})) + \mathbf{Y}_{i,l-1}, \\
    \mathbf{Y}_{i,l} = \mathrm{MLP}_l(\mathrm{LN}(\mathbf{Y}'_{i,l-1})) + \mathbf{Y}'_{i,l-1}, \\
    \mathbf{U}_i = (\mathbf{W}_\mathrm{u} \mathrm{LN}(\mathbf{Y}_{i,L})^\mathsf{T} + \mathbf{B}_\mathrm{u})^\mathsf{T}, \\
    \forall i \in \{1,2\}, \, l \in \{1, 2, \ldots, L\}, \notag
\end{gather}

\noindent where $\mathbf{c} \in \mathbb{R}^D, \mathbf{W}_\mathrm{e} \in \mathbb{R}^{D \times (P^2 \cdot C)}$, $\mathbf{B}_\mathrm{e} \in \mathbb{R}^{D \times N}$, $\mathbf{P}_\mathrm{e} \in \mathbb{R}^{(N+1) \times D}, \mathbf{W}_\mathrm{u} \in \mathbb{R}^{D' \times D}$, $\mathbf{B}_\mathrm{u} \in \mathbb{R}^{D' \times (N+1)}$, $[\, \cdot \, , \, \cdot \,]$ denotes the horizontal concatenation of matrices, and $L$ represents the number of Transformer blocks in the siamese encoder.

Subsequently, we replicate the trainable mask token $\mathbf{t}$ $\lvert \mathcal{P} \rvert$ times to create a matrix. This matrix is then incorporated into $\mathbf{U}_2$, and the consolidation of $\mathbf{U}_1$ and $\mathbf{U}_2$ is achieved through the following equations:
\begin{gather}
  \mathbf{T} = \mathrm{Repeat}(\mathbf{t}^\mathsf{T}, \lvert \mathcal{P} \lvert), \\
  \begin{aligned}
    \mathbf{U} = [\mathbf{U}_1, \quad & \mathrm{Convert}(\mathbf{U}_2, \mathcal{P}, N+1) \\
    + & \mathrm{Convert}(\mathbf{T}, \mathcal{P}^\complement, N+1)],
  \end{aligned}
\end{gather}

\noindent where $\mathcal{P}$ denotes the set of indices for the masked patches in the image, and $\lvert \, \cdot \, \rvert$ denotes the cardinality of the set.

\textbf{Reparameterization.} The features extracted by the siamese encoder traverse through the reparameterization layer, where the latent space is generated using a Gaussian distribution, enhancing the model's ability to produce varied and meaningful representations. From a mathematical standpoint, the reparameterization layer functions as follows:
\begin{gather}
  \mathbf{M} = (\mathbf{W}_\mathrm{m} \mathbf{U}^\mathsf{T} + \mathbf{B}_\mathrm{m})^\mathsf{T}, \\
  \mathbf{S} = (\mathbf{W}_\mathrm{s} \mathbf{U}^\mathsf{T} + \mathbf{B}_\mathrm{s})^\mathsf{T}, \\
  \mathbf{Z} = \mathbf{M} + \mathbf{S} \odot \mathbf{E},
\end{gather}

\noindent where $\mathbf{W}_\mathrm{m}, \mathbf{W}_\mathrm{s} \in \mathbb{R}^{D' \times 2D'}$, $\mathbf{B}_\mathrm{m}, \mathbf{B}_\mathrm{s} \in \mathbb{R}^{D' \times (N+1)}$, $\mathbf{E} \sim \mathcal{MN}_{(N+1) \times D'}(\mathbf{0}, \mathbf{I}, \mathbf{I})$, $\odot$ denotes the Hadamard product, and $\mathbf{Z}$ represents the latent matrix.

\textbf{Decoder.} The decoder in our framework is implemented as another vanilla ViT \cite{dosovitskiy2020image}. The decoder's core objective is to generate predictions for individual patches in pixel space, with the ultimate goal of reconstructing the initially missing content. The reconstruction operation is succinctly expressed through the following mathematical formulation:

\begin{gather}
  \mathbf{V}_{0} = (\mathbf{W}_\mathrm{d}[\mathbf{Z}, \mathbf{U}_1]^\mathsf{T} + \mathbf{B}_\mathrm{d})^\mathsf{T} + \mathbf{P}_\mathrm{d}, \\
  \mathbf{V}'_l = \mathrm{MSA}'_l(\mathrm{LN}(\mathbf{V}_{l-1})) + \mathbf{V}_{l-1}, \\
  \mathbf{V}_l = \mathrm{MLP}'_l(\mathrm{LN}(\mathbf{V}'_{l-1})) + \mathbf{V}'_{l-1}, \\
  \mathbf{O} = (\mathbf{W}_\mathrm{o} \mathrm{LN}(\mathbf{V}_{L'})^\mathsf{T} + \mathbf{B}_\mathrm{o})^\mathsf{T}, \\
  \forall l \in \{1, 2, \ldots, L'\}, \notag
\end{gather}

\noindent where $\mathbf{W}_\mathrm{d} \in \mathbb{R}^{D' \times 2D'}$, $\mathbf{B}_\mathrm{d} \in \mathbb{R}^{D' \times (N+1)}$, $\mathbf{P}_\mathrm{d} \in \mathbb{R}^{(N+1) \times D'}$, $\mathbf{W}_\mathrm{o} \in \mathbb{R}^{D' \times (P^2 \cdot C)}$, $\mathbf{B}_\mathrm{o} \in \mathbb{R}^{(P^2 \cdot C) \times (N+1)}$, and $L'$ represents the number of Transformer blocks in the decoder.

Finally, we integrate the predicted masked patches with the unmasked patches from the original image using the following operation:
\begin{equation}
  \mathbf{G} = \mathrm{Convert}([\mathbf{0}^\mathsf{T}; \mathbf{X}_2], \mathcal{P}, N) + \mathrm{Convert}(\mathbf{O}, \mathcal{P}^\complement, N),
\end{equation}

\noindent where $[\, \cdot \, ; \, \cdot \,]$ denotes the vertical concatenation of matrices.

\textbf{Loss Function.} Inspired by $\beta$-VAE \cite{higgins2016beta}, we model the prior as an isotropic unit Gaussian $\mathcal{MN}(\mathbf{0}, \mathbf{I}, \mathbf{I})$, leading to the formulation of the constrained optimization problem:
\begin{equation}
  \begin{gathered}
    \mathrm{max}_{\phi, \theta} \mathbb{E}_{\mathbf{X}_1, \mathbf{X}_2 \sim \mathcal{D}}[\mathbb{E}_{q_\phi(\mathbf{Z} \mid \mathbf{X}_1, \mathbf{X}_2)} \log p_\theta(\mathbf{R} \mid \mathbf{Z})], \\
    \mathrm{s.t.} \, D_\mathrm{KL}(q_\phi(\mathbf{Z} \mid \mathbf{X}_1, \mathbf{X}_2) \| p(\mathbf{Z})) \leq \epsilon,
  \end{gathered}
\end{equation}

We reformulate it as a Lagrangian under the KKT conditions \cite{caron2021emerging}:
\begin{equation}
  \label{eq:KKT-condition}
  \begin{gathered}
    \mathcal{F}(\theta, \phi, \beta; \mathbf{X}_1, \mathbf{X}_2, \mathbf{R}) = \mathbb{E}_{q_\phi(\mathbf{Z} \mid \mathbf{X}_1, \mathbf{X}_2)} \log p_\theta(\mathbf{R} \mid \mathbf{Z}) \\
    - \beta(D_\mathrm{KL}(q_\phi(\mathbf{Z} \mid \mathbf{X}_1, \mathbf{X}_2) \| p(\mathbf{Z})) - \epsilon),
  \end{gathered}
\end{equation}

As $\epsilon$ is a constant, it is disregarded in the optimization. Our training strategy for SiamMCVAE involves the formulation of a comprehensive loss function that combines both a reconstruction loss ($\mathcal{L}_{\mathrm{r}}$) and a KL divergence loss ($\mathcal{L}_{\mathrm{KL}}$). The structure of the loss function is articulated as follows:
\begin{equation}
  \label{eq:overall_loss}
  \mathcal{L} = \mathcal{L}_\mathrm{r} + \beta \cdot \mathcal{L}_\mathrm{KL}
\end{equation}

\noindent where $\beta$ is a hyperparameter that controls the trade-off between the two components.

The reconstruction loss, integral to our model's training, quantifies the disparity between the original and reconstructed data and is formulated as follows:
\begin{equation}
  \label{eq:reconstruction_loss}
  \mathcal{L}_{\mathrm{r}} = \frac{1}{P^2C \lvert \mathcal{P} \rvert} \| \mathbf{G} - \mathbf{R} \|^2_\mathrm{F}
\end{equation}

\noindent where $\mathbf{R}$ represents the patchified target image. 

The KL divergence loss, which measures the dissimilarity between the learned latent distribution and a chosen prior distribution, is given by:
\begin{equation}
  \label{eq:KL_loss}
  \mathcal{L}_{\mathrm{KL}} = \frac{\| \mathbf{M} \|_\mathrm{F}^2 + \| \mathbf{S} \|_\mathrm{F}^2 -\sum_{i=1}^{N+1} \sum_{j=1}^{D'} \log \mathbf{S}_{ij}}{2(N+1)D'} - \frac{1}{2}
\end{equation}

\noindent where $\| \cdot \|_\mathrm{F}$ denotes the Frobenius norm.

The overall loss function optimizes the model to minimize the reconstruction error while encouraging the latent distribution to be close to the chosen prior. This combination ensures that the SiamMCVAE effectively reconstructs the lost content in video frames.

%% file: sec/4_exper.tex
\section{Experiments}
In this section, we embark on a comprehensive evaluation of the performance of our SiamMCVAE model, juxtaposing it against established state-of-the-art methodologies. This systematic assessment seeks to shed light on the model's capabilities and its potential to address real-world challenges.

\subsection{Experiment Setup}
\textbf{Dataset.} Our experiments are conducted on the extensive BDD100K dataset \cite{yu2020bdd100k}, renowned for its diverse range of driving scenarios. Encompassing a rich collection of images and videos, BDD100K provides a comprehensive array of scenarios and environments commonly encountered on roadways \cite{cui2022dg}. For our evaluation of the SiamMCVAE model, we meticulously select a curated subset of video sequences, ensuring a representative sampling across diverse real-world scenarios and challenges.

\textbf{Masking.} Our masking strategy involves the deliberate occlusion of a segment within one frame of a paired set of images, while the other frame remains unaltered. This deliberate masking of a portion of the image serves as a surrogate for scenarios in which partial data loss or image corruption occurs in dynamic video sequences.

\textbf{Evaluation metrics.} Our evaluation strategy employs a meticulous selection of metrics designed to thoroughly assess the quality of the restored frames in comparison to the ground truth. In addition to the conventional Mean Squared Error (MSE) and Mean Absolute Error (MAE), we leverage the Peak Signal-to-Noise Ratio (PSNR), a well-established measure offering valuable insights into the model's precision in capturing fine details and minimizing differences in pixel values.

For a thorough evaluation, we incorporate advanced metrics, notably the Structural Similarity Index (SSIM) \cite{wang2004image} and the Feature-based Similarity Index (FSIM) \cite{zhang2011fsim}. These sophisticated indices augment our assessment by providing a nuanced perspective on the model's performance. By scrutinizing the structural similarity between the restored and ground truth frames, encompassing considerations such as luminance, contrast, and structure, these metrics go beyond pixel-level accuracy. They offer valuable insights into the model's adeptness in preserving the overall structural coherence and visual fidelity of the restored frames.

The orchestration of this ensemble of metrics in our evaluation provides a nuanced and comprehensive view of our model's prowess in video frame restoration.

\subsection{Comparison with Prior Work}
We systematically conduct a comprehensive performance analysis, pitting our SiamMCVAE model against baseline methods, including MAE \cite{he2022masked}, MAE-ST \cite{feichtenhofer2022masked}, and VideoMAE \cite{tong2022videomae}, within the domain of video frame restoration. Our meticulous evaluation focuses on a masking ratio of 75\%, representing a scenario characterized by moderate data degradation. The outcomes specific to this masking ratio are concisely presented in \Cref{table-restoration-performance}, offering valuable insights into the comparative efficacy of our model and established baselines.

\begin{table*}[t]
  \centering
  \begin{tabular}{|c|c|c|c|c|c|c|}
    \hline
    Method & Backbone & MSE & MAE & PSNR & SSIM & FSIM\\
    \hline
    MAE \cite{he2022masked} & ViT-B & 197.37 & 6.99 & 25.80 & 0.800 & 0.670 \\
    MAE-ST \cite{feichtenhofer2022masked} & ViT-B & 258.51 & 8.11 & 24.70 & 0.741 & 0.638 \\
    VideoMAE \cite{tong2022videomae} & ViT-B & 198.00 & 6.97 & 25.80 & 0.798 & 0.669 \\
    \hline
    MAE \cite{he2022masked} & ViT-L & 146.63 & 5.95 & 27.10 & 0.837 & 0.700 \\
    MAE-ST \cite{feichtenhofer2022masked} & ViT-L & 221.69 & 7.58 & 25.34 & 0.758 & 0.651 \\
    VideoMAE \cite{tong2022videomae} & ViT-L & 133.83 & 5.61 & 27.51 & 0.838 & 0.708 \\
    \hline
    \textbf{SiamMCVAE} (ours) & SiamViT & \textbf{123.01} & \textbf{5.49} & \textbf{27.90} & \textbf{0.841} & \textbf{0.712} \\
    \hline
  \end{tabular}
  \caption{\textbf{Performance comparison with prior work} on restoration metrics at a 75\% masking ratio. Our proposed method, SiamMCVAE outperforms the existing approaches across various metrics, showcasing its superior ability in restoring missing information in video frames.}
  \label{table-restoration-performance}
\end{table*}

It is noteworthy that our SiamMCVAE model consistently outperforms the baseline methods across a spectrum of comprehensive evaluation metrics, namely, MAE, MSE, PSNR, SSIM, and FSIM. The prominent superiority observed in these metrics emphasizes the model's exceptional proficiency in minimizing both subtle and substantial reconstruction errors. Consequently, SiamMCVAE stands out as a benchmark in the field of video frame restoration.

These results underscore the efficacy of our SiamMCVAE model, not only in mitigating the effects of data degradation but also in surpassing established state-of-the-art methods in the field of video frame restoration. The capacity to excel in such a challenging scenario further solidifies the model's potential for real-world applications where data integrity may be compromised.

\subsection{Model Robustness}
\begin{figure}[t]
  \centering
  \begin{tikzpicture}
    \begin{axis}[
      width=\linewidth,
      xlabel={Masking ratio (\%)},
      ylabel={MSE},
      xtick={45,60,75,90},
      legend pos=north west
    ]
    
    \addplot[blue, solid, mark=o] coordinates {
      (45,83.13)
      (60,127.53)
      (75,197.37)
      (90,408.77)
    };
    \addlegendentry{MAE}
    
    \addplot[orange, dash pattern=on 3pt off 3pt on 1pt off 3pt, thick, mark=star] coordinates {
      (45,120)
      (60,181.06)
      (75,258.51)
      (90,322.38)
    };
    \addlegendentry{MAE-ST}
    
    \addplot[violet, dash pattern=on 4pt off 2pt, mark=+] coordinates {
      (45,78.71)
      (60,122.28)
      (75,198)
      (90,379.05)
    };
    \addlegendentry{VideoMAE}
    
    \addplot[red, solid, thick, mark=diamond] coordinates {
      (45,74.81)
      (60,104.18)
      (75,139.22)
      (90,218.94)
    };
    \addlegendentry{SiamMCVAE}
    
    \end{axis}
  \end{tikzpicture}
  \caption{Performance comparison of different models across varying masking ratios. In the face of increasing masking ratios, SiamMCVAE consistently outperforms other models, showcasing its remarkable resilience and effectiveness in restoring missing information within video frames.}
  \label{fig:masking-ratio-performance}
\end{figure}
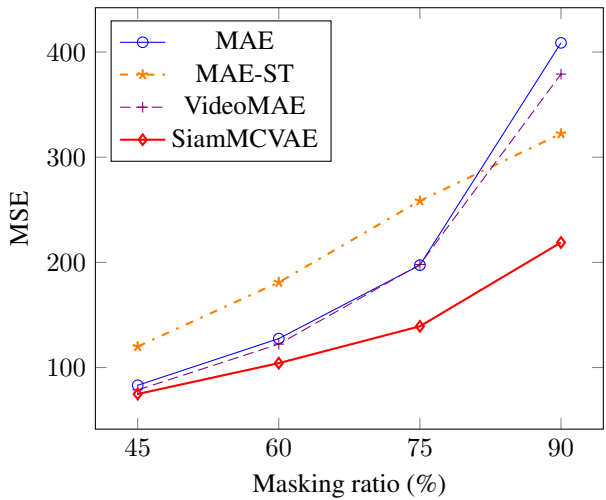

Through extensive experimentation conducted on diverse driving scenarios extracted from the dataset, we employ a spectrum of masking ratios spanning from 45\% to 90\%, encompassing a diverse range of damage severity scenarios. This intentional variation in mask coverage enables us to perform a nuanced and thorough assessment of our model's proficiency in restoring video frames across a spectrum of degradation conditions. The outcomes depicted in \Cref{fig:masking-ratio-performance} underscore the remarkable superiority of SiamMCVAE over other models in the face of diverse levels of data degradation. This pronounced ascendancy becomes particularly conspicuous when the masking ratio attains higher thresholds.

\begin{figure}[t]
  \centering
  \begin{tikzpicture}
    \begin{axis}[
      width=\linewidth,
      xlabel={Frame gap},
      ylabel={MSE},
      xtick={24,36,48,60},
      legend style={fill=none},
      legend pos=north west
    ]
    
    \addplot[orange, dash pattern=on 3pt off 3pt on 1pt off 3pt, thick, mark=star] coordinates {
      (24,258.51)
      (36,442.07)
      (48,475.11)
      (60,334.24)
    };
    \addlegendentry{MAE-ST}
    
    \addplot[violet, dash pattern=on 4pt off 2pt, mark=+] coordinates {
      (24,198)
      (36,201.3)
      (48,199.26)
      (60,199.49)
    };
    \addlegendentry{VideoMAE}
    
    \addplot[red, solid, thick, mark=diamond] coordinates {
      (24,139.22)
      (36,150.69)
      (48,153.81)
      (60,156.29)
    };
    \addlegendentry{SiamMCVAE}
    
    \end{axis}
  \end{tikzpicture}
  \caption{Performance comparison across different frame gaps. Notably, the SiamMCVAE consistently outperforms both MAE-ST and VideoMAE.}
  \label{fig:frame-gap-performance}
\end{figure}
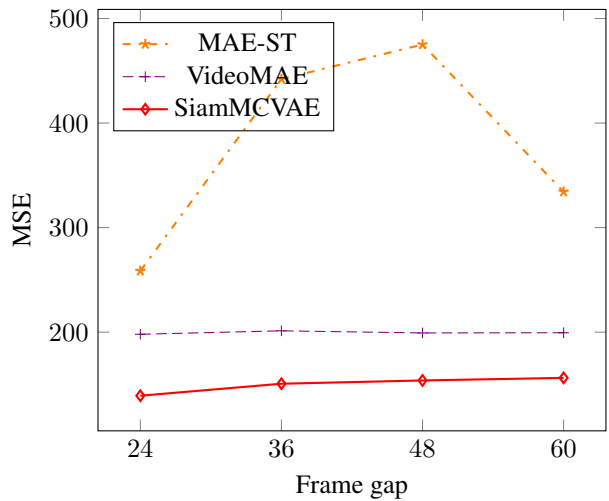

Furthermore, we evaluate the performance of various models across different frame gap scenarios, illustrated in \Cref{fig:frame-gap-performance}. What stands out conspicuously is the persistent dominance of SiamMCVAE, regardless of the frame gap setting. This sustained advantage serves as a testament to the model's exceptional adaptability and robustness.

\subsection{Qualitative Analysis}
\begin{figure*}[t]
  \centering
  \begin{subfigure}{\textwidth}
    \includegraphics[width=\textwidth]{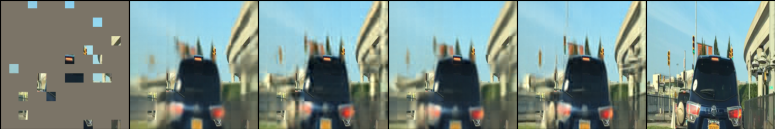}
  \end{subfigure}
  \begin{subfigure}{\textwidth}
    \includegraphics[width=\textwidth]{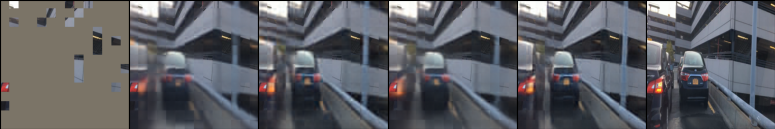}
  \end{subfigure}
  \begin{subfigure}{\textwidth}
    \includegraphics[width=\textwidth]{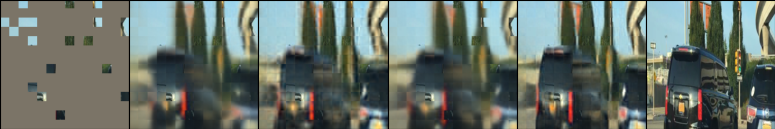}
  \end{subfigure}
  \begin{subfigure}{\textwidth}
    \includegraphics[width=\textwidth]{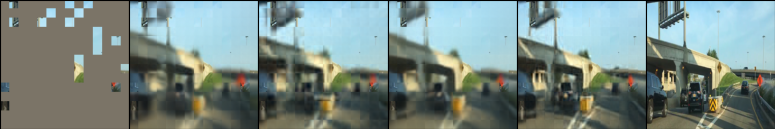}
  \end{subfigure}
  \caption{\textbf{Comparative visualization} of model outputs at a 90\% masking ratio. In the first column, masked video frames are depicted, while the subsequent columns showcase outputs from various models, including MAE \cite{he2022masked}, MAE-ST \cite{feichtenhofer2022masked}, VideoMAE \cite{tong2022videomae}, and our SiamMCVAE, arranged from left to right. The rightmost column features the unaltered ground truth frames.}
  \label{fig:model-comparison}
\end{figure*}

In our pursuit of a comprehensive evaluation, we delve into the qualitative facets of model performance. To this end, we embark on a visual exploration of model outputs when faced with masked video frames. The resulting visualizations, exemplified in \Cref{fig:model-comparison}, offer a nuanced perspective on the reconstruction capabilities across various models. The visual comparisons distinctly reveal the superior performance of SiamMCVAE in terms of the quality of restored images when compared to alternative models.

\subsection{Ablation Studies}
\textbf{Attention kernel.} In-depth exploration of attention kernels is crucial for understanding their nuanced impact on the efficacy of our SiamMCVAE model. We systematically assess the performance by comparing the adaptive attention kernel with established counterparts such as Standard Attention \cite{vaswani2017attention}, Flash Attention \cite{dao2022flashattention}, and Memory-Efficient Attention \cite{jeevan2022resource}. The discerning outcomes of this comparative analysis are succinctly summarized in \Cref{table-attention-analysis}, offering a comprehensive perspective on how different attention kernels influence the overall performance of the model.

\begin{table}[t]
  \centering
  \begin{tabular}{|c|c|c|c|c|c|}
    \hline
    Kernel & MSE & MAE & PSNR & SSIM & FSIM \\
    \hline
    SA \cite{vaswani2017attention} & 160.29 & 6.32 & 26.75 & 0.806 & 0.687 \\
    FA \cite{dao2022flashattention} & 143.56 & 5.96 & 27.25 & 0.456 & 0.697 \\
    MEA \cite{jeevan2022resource} & 156.05 & 6.23 & 26.87 & 0.810 & 0.689 \\
    Adaptive & \textbf{123.01} & \textbf{5.49} & \textbf{27.90} & \textbf{0.841} & \textbf{0.712} \\
    \hline
  \end{tabular}
  \caption{Comparison of Standard Attention (SA) \cite{vaswani2017attention}, Flash Attention (FA) \cite{dao2022flashattention}, Memory-Efficient Attention (MEA) \cite{jeevan2022resource}, and the adaptive attention kernel on SiamMCVAE Performance.}
  \label{table-attention-analysis}
\end{table}

\textbf{Reparameterization layer.} To gain deeper insights into the inner workings of our SiamMCVAE architecture, we conducted a meticulous comparative analysis between models with and without the reparameterization layer. The compelling results, detailed in \Cref{table-reparameterization-analysis}, underscore the substantial performance improvement achieved through the incorporation of the reparameterization layer. Evident from the reduced MSE and MAE, as well as the elevated PSNR, SSIM, and FSIM scores, this analysis emphasizes the pivotal role of reparameterization in enhancing the model's overall restoration capabilities.

\begin{table}[t]
  \centering
  \begin{tabular}{|c|c|c|c|c|c|}
    \hline
    Reparam. & MSE & MAE & PSNR & SSIM & FSIM \\
    \hline
    \texttimes & 174.86 & 6.63 & 26.35 & 0.792 & 0.676 \\
    \checkmark & \textbf{123.01} & \textbf{5.49} & \textbf{27.90} & \textbf{0.841} & \textbf{0.712} \\
    \hline
  \end{tabular}
  \caption{Comparison of SiamMCVAE performance: without reparameterization (\texttimes) vs. with reparameterization (\checkmark).}
  \label{table-reparameterization-analysis}
\end{table}

\textbf{Lagrange multiplier.} Within the intricacies of our SiamMCVAE model, we scrutinize the impact of the Lagrange multiplier, denoted as $\beta$. As elucidated in \Cref{table-beta-performance}, we conduct a thorough analysis of the model's performance across varying $\beta$ values. This examination provides nuanced insights into the delicate interplay between regularization strength and restoration efficacy. The results underscore the importance of meticulous tuning of $\beta$ to strike a balance, ensuring optimal expressiveness while preserving crucial visual details. Notably, the analysis identifies $\beta = 0.2$ as the optimal value, showcasing superior performance across multiple evaluation metrics.

\begin{table}[t]
  \centering
  \begin{tabular}{|c|c|c|c|c|c|}
    \hline
    $\beta$ & MSE & MAE & PSNR & SSIM & FSIM \\
    \hline
    0.1 & 151.42 & 6.14 & 26.98 & 0.812 & 0.691\\
    0.2 & \textbf{123.01} & \textbf{5.49} & \textbf{27.90} & \textbf{0.841} & \textbf{0.712} \\
    0.25 & 139.22 & 5.89 & 27.36 & 0.825 & 0.700 \\
    0.5 & 172.02 & 6.57 & 26.43 & 0.809 & 0.678\\
    1 & 192.74 & 7.01 & 25.90 & 0.777 & 0.666 \\
    \hline
  \end{tabular}
  \caption{Impact of Lagrange Multiplier ($\beta$) on SiamMCVAE Performance. The results demonstrate the model's sensitivity to the choice of $\beta$. Notably, the highlighted values indicate the superior performance achieved with a $\beta$ value of 0.2.}
  \label{table-beta-performance}
\end{table}

%% file: sec/5_discu.tex
\section{Discussion}
The SiamMCVAE model takes a prominent position in the field of video frame restoration, showcasing remarkable efficacy in scenarios characterized by substantial information loss. Through the synergistic integration of the innovative SiamViT and variational inference, our model excels in the task of restoration, solidifying its status as a state-of-the-art solution.

Through extensive experimentation conducted on diverse driving scenarios extracted from the BDD100K dataset \cite{yu2020bdd100k}, SiamMCVAE consistently outshines its other models across various mask ratios and diverse frame gap settings. This resounding success underscores its remarkable adaptability, demonstrating superior performance even in challenging conditions. The robustness of SiamMCVAE can be attributed to careful design considerations, including the strategic integration of SiamViT and the judicious application of variational techniques. These elements collectively contribute to the model's adaptability, positioning it as a resilient and superior solution capable of addressing a spectrum of challenges in video frame restoration.

Our exhaustive ablation study, meticulously scrutinizing the influence of crucial components, illuminates the efficacy of the SiamMCVAE model's design. We explicitly investigate the roles played by attention mechanisms, the reparameterization layer, and the Lagrange multiplier $\beta$. This in-depth analysis quantifies the distinct contributions of these elements, offering a profound insight into the nuanced design choices that form the bedrock of our model's success.

%% file: sec/6_concl.tex
\section{Conclusion}
The successful fusion of siamese architectures with advanced vision transformers, exemplified by SiamMCVAE, presents a significant leap forward in the domain of video frame restoration under masked scenarios. The incorporation of variational principles adds another layer of innovation, enhancing the model's capacity to generate diverse and meaningful representations. Beyond the immediate context of video frame restoration, our work highlights the broader potential of synergizing siamese encoders with state-of-the-art vision transformers \cite{dosovitskiy2020image} for generative purpose. SiamMCVAE not only pushes the boundaries of restoration capability but also sets a precedent for the integration of these advanced architectures, including variational techniques, in addressing real-world challenges within the expansive field of computer vision.